\documentclass[11pt]{article}
\usepackage{latexsym}
\usepackage{amsfonts}
\usepackage{graphicx}
\usepackage{epstopdf}
\usepackage{xcolor}
\usepackage[sort]{natbib}
\usepackage{lineno}
\newcommand{\coln}{\hspace*{-6pt}{\bf :}}
\newtheorem{theorem}{Theorem}
\newtheorem{lemma}{Lemma}
\newtheorem{corol}{Corollary}
\newtheorem{property}{Property}

\newcommand{\Theorem}[1]{\begin{theorem}\coln ~#1 \end{theorem}}

\newcommand{\proof}[1]{\noindent{\bf Proof:~}#1\hfill ~$\Box$}
\newcommand{\be}{\begin{equation}}
\newcommand{\ee}{\end{equation}}

\newcommand{\narrow}{\setlength{\itemsep}{1pt}
	\setlength{\parskip}{0pt}\setlength{\parsep}{0pt}}

\begin{document}
	\title{\bf The Importance of Good Starting Solutions in the Minimum Sum of Squares Clustering Problem}
	\author{Pawel Kalczynski\\
		Steven G. Mihaylo College of Business and Economics\\
		California State University-Fullerton\\
		Fullerton, CA 92834.\\e-mail: pkalczynski@fullerton.edu\and	
		Jack Brimberg\\Department of Mathematics and Computer Science\\The Royal Military College of Canada\\ Kingston, ON Canada.\\ e-mail: Jack.Brimberg@rmc.ca\and Zvi Drezner\footnote{Corresponding author}\\
		Steven G. Mihaylo College of Business and Economics\\
		California State University-Fullerton\\
		Fullerton, CA 92834.\\e-mail: zdrezner@fullerton.edu
	}
	\date{}
	\maketitle
	
	\begin{abstract}
The clustering problem has many applications in Machine Learning, Operations Research, and Statistics. We propose three algorithms to create starting solutions for improvement algorithms for this problem. We test the algorithms on 72 instances that were investigated in the literature. Forty eight of them are relatively easy to solve and we found the best known solution many times for all of them. Twenty four medium and large size instances are more challenging. We found five new best known solutions and matched the best known solution for 18 of the remaining 19 instances.		~
	\end{abstract}

	\renewcommand{\baselinestretch}{1.5}
	\renewcommand{\arraystretch}{0.67}
	\large
	\normalsize
	
	\section{Introduction}
	
	{\color{black}We present three new approaches 
	that may be used to generate good starting solutions to clustering problems and then we} show that these
	starting solutions vastly improve the performance of popular local searches (improvement algorithms)
	used in clustering such as k-means \citep{L57,HW79,M67a}. In several benchmark cases, we were even able to improve 
	on 
	best-known results.

A set of $n$ points  is given.	The problem is to find $k$ cluster centers.  Each point is assigned to its closest center. The objective is to minimize the total sum of squares of the distances to the closest cluster center. Let $d_{i}(X_j)$ be the Euclidean distance between given point $i$ and center of cluster $j$ located at $X_j$. The points are represented in $d$-dimensional space. The vector of unknown centers is ${\bf X}=\{X_1,\ldots,X_k\}$, and thus, the objective function to be minimized is:
	\begin{equation}\label{probK}
	F({\bf X})=\sum\limits_{i=1}^n \min\limits_{1\le j\le k}\left\{d_{i}^2(X_j)\right\}
	\end{equation}
	
Recent publications on clustering problems minimizing the sum of squared distances between all given points and their cluster's center are  \cite{A09}, {\color{black}\cite{AHL12},} \cite{BODX15}, \cite{PABM19} {\color{black} and \cite{GV19}.
		
\citet{GV19} recently suggested a complex hybrid genetic algorithm for the solution of the problem which obtains best reported results to date. Our approach applies a multi-start local search using a set of specially designed high quality starting solutions. Despite the relative simplicity of our method, \citet{GV19} obtained a better solution in only one instance out of the 72 instances tested here. These results confirm in a dramatic way the importance of `good' starting solutions for the minimum sum of squares clustering problem. This also corroborates the `less is more' philosophy adopted in some recent papers on heuristic design \citep[e.g., ][]{MTU16,BMT17}. In this case we find that a simple local search from diverse starting solutions of good quality can be as powerful as a sophisticated meta-heuristic based method.}
	
This problem has similarities to the $p$-median problem  \citep{Das95,KH79med,DM15}, also called the multi-source Weber problem  \citep{BHMT00,KS72}. The objective of the $p$-median problem is based on the distances rather than the squared distances and each point (or customer) has a positive weight, while in formulation (\ref{probK}) all weights are equal to 1.
	
	\subsection{\label{widely}Widely Used Algorithms}
	
	{\color{black}Algorithms similar to k-means include \citet{S85,L57,Forgy1965,David2007,BMV12,M67a,HW79}}.
	
	\begin{description}

		\item[{\color{black}Algorithm of Lloyd \citep{L57}:}]
		
		The Lloyd algorithm is an iterative improvement heuristic which consists of repeating two steps, assignment and update, until there are no more changes in the assignment of points to clusters or the maximum number of iterations is reached. The initial set of $k$ centers $X=\{X_1,X_2,...,X_k\}$ and maximum number of iterations are  inputs to the algorithm. The initial set of centers can be determined by random assignment of points to $k$ groups (as in the original Lloyd algorithm) or by an initialization method such as Forgy \citep{Forgy1965}, k-means++ \citep{David2007}, or scalable k-means++ \citep{BMV12}.
		{\color{black}In k-means++, the starting solution consists of a subset of original data points selected by a seeding algorithm, which spreads the initial centroids based on their squared distances from the first centroid.}
		
		In the assignment step, the current set of centers is used to determine the Voronoi cells \citep{V08,OKSC00}, and each point $x_i$ is assigned to the nearest center in terms of the squared Euclidean distance:
		
		\begin{equation}\label{voronoi}
		V_\ell=\{x_i:d^2_i(X_\ell) \le d^2_i(X_j),	\forall j, 1\le j \le k \}, \ell=1,\ldots k
		\end{equation}
		in such a manner that each point is assigned to exactly one Voronoi cell.
		Next, in the update step, the new centers are determined by finding the centroid of points in each cell:
		\begin{equation}\label{centroids}
		X^\prime_\ell=\frac{1}{|V_\ell|}\sum_{x_i \in V_\ell}x_i.
		\end{equation}
		
		The two steps are repeated in the original order until there is no change in the assignment of points to clusters.

		This algorithm is very similar to the location-allocation procedure (ALT) proposed by \citet{Co63,Co64} for the solution of the Euclidean p-median problem. However, for the squared Euclidean objective, the solution is expressed as a simple formula (\ref{centroids}), while for the weighted Euclidean distances (not squared), the center of the cluster requires a special algorithm such as \citet{W36,Dr13} which requires longer run times.

		\item[{\color{black}Algorithm of MacQueen \citep{M67a}:}]
		The initialization of the MacQueen algorithm is identical to Lloyd, i.e., the initial centers are used to assign points to clusters.
		
		In the iteration phase, the MacQueen heuristic reassigns only those points which are nearer a center different from the one they are currently assigned to. Only the centers for the original and new clusters are recalculated after the change which improves the efficiency of the heuristic as compared to Lloyd.
		
		The improvement step is repeated until there is no change in the assignment of points to clusters.

		\item[{\color{black}Algorithm of Hartigan-Wong \citep{HW79}:}]
		
		The initialization of the Hartigan-Wong algorithm is identical to MacQueen and Lloyd and points are assigned to centers using the Voronoi's method. However, the improvement step uses the within-group sum of squared Euclidean distances $\sum\limits_{x_i \in V_\ell} d^2_i(X_\ell)$, where $V_\ell$ is a cluster centered at $X_\ell$. 
		
		Specifically, for each previously-updated cluster $V_j$, the point $x_i \in V_j$ is reassigned to $V_\ell$ ($\ell \ne j$) if such a reassignment reduces the total within-group sum of squared distances for all clusters.
		
		The improvement step is repeated until there is no change in the assignment of points to clusters.
	\end{description}
	
	The paper is organized as follows. In Section \ref{sec2} we discuss the execution of basic operations applied in the paper. In Section \ref{sec3} the algorithms designed in this paper are described. In Section \ref{refine} a new improvement algorithm is detailed. In Section \ref{sec4} we report the results of the computational experiments, and we conclude the paper in Section \ref{sec5}.
	
	\section{\label{sec2}Preliminary Analysis: Basic Operations}
	In this paper we apply three basic operations: adding a point to a cluster, removing a point from a cluster, and combining two clusters into one. The objective function is separable to individual dimensions and thus we show the calculation in one dimension. In $d$ dimensions the change in the value of the objective function is the sum of the changes in each dimension.
	
	Let a cluster consist of $m$ points $x_i,i=1,\ldots,m$ with a mean {\color{black}$\bar x_m=\frac1m\sum\limits_{i=1}^mx_i$. Since $\sum\limits_{i=1}^mx_i=m\bar x_m$,}
the objective function is:
	$$F_m=\sum\limits_{i=1}^m (x_i-\bar x_m)^2~=~{\color{black}\sum\limits_{i=1}^mx_i^2~-2\bar x_m\sum\limits_{i=1}^mx_i+~m\bar x_m^2~=~}\sum\limits_{i=1}^mx_i^2~-~m\bar x_m^2.$$
	
	\subsection{Adding a point to a cluster}
	\Theorem{\label{Th1}When a point $x_{m+1}$ is added to the cluster, the objective function is increased by $\frac{m}{m+1}(x_{m+1}-\bar x_m)^2$.}
	\proof{	
		The new center is at
		$$\bar x_{m+1}=\frac{m\bar x_m+x_{m+1}}{m+1}=\bar x_{m}+\frac{x_{m+1}-\bar x_m}{m+1}\equiv \bar x_m+\Delta x_m
		$$
		
		We get
		\begin{eqnarray}
		F_{m+1}&=&F_m+x_{m+1}^2 +m\bar x_m^2-(m+1)\bar x_{m+1}^2
		\nonumber\\&=&F_m+\left[\bar x_{m}+(m+1)\Delta x_m\right]^2+m\bar x_m^2-(m+1)\left[\bar x_{m}+\Delta x_m\right]^2
		=F_m+m(m+1)\left[\Delta x_m\right]^2\nonumber
		\end{eqnarray}
		It can be written as
		\begin{equation}\label{add}
		F_{m+1}=F_m+\frac{m}{m+1}(x_{m+1}-\bar x_m)^2
		\end{equation}
		which proves the theorem.
	}
	
	\subsection{Removing a point from a cluster}
	
	\Theorem{\label{Th2}Suppose that $x_m$ is removed from a cluster. The reduction in the value of the objective function is: $\frac{m}{m-1}\left(x_{m}-\bar x_m\right)^2$.}
	\proof{The new center is at
		$$\bar x_{m-1}=\frac{m\bar x_m-x_{m}}{m-1}=\bar x_{m}-\frac{x_{m}-\bar x_m}{m-1}
		$$
		By equation (\ref{add}) for $m-1$
		\begin{eqnarray}\label{sub}
		F_{m-1}&=&F_m-\frac{m-1}{m}(x_{m}-\bar x_{m-1})^2\nonumber\\&=&F_m-\frac{m-1}{m}\left(x_{m}-\frac{m\bar x_m-x_{m}}{m-1}\right)^2
		=F_m-\frac{m}{m-1}\left(x_{m}-\bar x_m\right)^2
		\end{eqnarray}
		which proves the theorem.
	}
	
	\subsection{Combining two clusters}
	\Theorem{\label{Th3}Two clusters of sizes $m_1$ and $m_2$ with centers at $\bar x_{m_1}$ and $\bar x_{m_2}$ are combined into one cluster of size $m_1+m_2$. The increase in the value of the objective function is $\frac{m_1m_2}{m_1+m_2}\left[\bar x_{m_1}- \bar x_{m_2}\right]^2$.} 
	\proof{
		The center of the combined cluster is at:
		\begin{equation}\label{case3}
		\bar x_{m_1+m_2}=\frac{m_1\bar x_{m_1}+m_2\bar x_{m_2}}{m_1+m_2}
		\end{equation}
		The objective function $F_{m_1+m_2}$ is:
		\begin{eqnarray}
		F_{m_1+m_2}&=&\sum\limits_{i=1}^{m_1+m_2}x_i^2~-~(m_1+m_2)\bar x_{m_1+m_2}^2
		\nonumber\\&=&F_{m_1}+F_{m_2}+m_1\bar x_{m_1}^2+m_2\bar x_{m_2}^2-(m_1+m_2)\bar x_{m_1+m_2}^2
		\nonumber\\&=&F_{m_1}+F_{m_2}+m_1\bar x_{m_1}^2+m_2\bar x_{m_2}^2-\frac{\left[m_1\bar x_{m_1}+m_2\bar x_{m_2}\right]^2}{m_1+m_2}
		\nonumber\\&=&F_{m_1}+F_{m_2}+\frac{m_1m_2\bar x_{m_1}^2+m_1m_2\bar x_{m_2}^2-2m_1m_2\bar x_{m_1}\bar x_{m_2}}{m_1+m_2}
		\nonumber\\&=&F_{m_1}+F_{m_2}+\frac{m_1m_2}{m_1+m_2}\left[\bar x_{m_1}- \bar x_{m_2}\right]^2\label{sum}
		\end{eqnarray}
		which proves the theorem.
	}
	\subsection{Multi-dimensional points}
	The following theorem  considers clustering of points in $d$ dimensions. \Theorem{\label{Th4}Once the clusters' centers and number of points in each cluster are saved in memory, the change in the value of the objective function by adding a point to a cluster, removing a point, or combining two clusters is calculated in $O(d)$.}
	\proof{By Theorems \ref{Th1}-\ref{Th3} the calculation in each dimension is done in $O(1)$ and thus the $d$-dimensional calculation is done in $O(d)$ as a sum of $d$ terms.}

	\section{\label{sec3}Finding a Starting Solution}
	
	We find initial sets of clusters that can serve as starting solutions for various improvement algorithms. The procedures are based on three algorithms proposed in recent papers for solving various multi-facility location problems, that can be easily extended to the clustering problem.
	
	The first two algorithms described below can be applied without a random component yielding one starting solution. We introduce randomness into the procedures so that the algorithms can be repeated many times in a multi-start heuristic approach and add some diversification to the search. The randomness idea follows the ``Greedy Randomized Adaptive Search Procedure" (GRASP) suggested by \citet{FR95}. {\color{black}It is a greedy approach but in each iteration the move is randomly selected by some rule, rather than always selecting the best one.} For each of the first two algorithms, a different GRASP approach is used and details are provided for each in the appropriate sub-section.

	\subsection{Merging \citep{BDMS12a}}
	
	The merging algorithm is based on the START algorithm presented in \cite{DBMS13,BDMS12a} for the solution of the planar $p$-median problem, also known as the multi-source Weber problem. {\color{black} The START algorithm begins} with $n$ clusters, each consisting of one given point. We evaluate combining pairs of clusters and combine the pair which increases the value of the objective function the least, thereby reducing the number of clusters by one. The process continues until $k$ clusters remain.
	
In order to randomly generate starting solutions to be used in a multi-start approach we used the following GRASP approach. We randomly select a pair of clusters within a specified factor, $\alpha>1$, of the minimum increase. {\color{black} For $\alpha=1$ the move with the minimum increase is always selected. When $\alpha$ increases, more moves, which are not close to the minimum, can be selected.} To simplify the procedure we follow the approach proposed in \cite{Dr10b}. Set $\Delta$ to a large number. The list of increases in the value of the objective function is scanned. If a new best increase $\Delta$ is found, update $\Delta$, select the pair of clusters, and set $r=1$. If an increase less than $\alpha\Delta$ is found, set $r=r+1$ and replace the selected pair with probability $\frac1r$.

	The basic merging algorithm is:
	
	\begin{enumerate}
		\item $n$ clusters are defined, each containing one point. Set $m_i=1$
		for  $i=1,\ldots,n$.
		\item Repeat the
		following until the number of clusters is reduced to $k$.
		\begin{enumerate}
			\narrow
			\item Find the pair $i<j$ for which the increase in the value of the objective function by Theorem \ref{Th3} is minimized (applying GRASP). If an increase of 0 is found, there is no need to continue the evaluation of pairs; skip to \ref{b}.
			\item \label{b} Combine the selected clusters $\{i,j\}$, find the new center by {\color{black}equation} (\ref{case3}) and replace $m_i$ by $m_i+m_j$.
			\item Remove clusters $i$ and $j$, and add the new cluster. The number of clusters is reduced by one.
		\end{enumerate}
	\end{enumerate}

	The complexity of this algorithm is $O(n^3d)$. Only the number of points in each cluster and their centers need to be kept in memory. The list of points belonging to each cluster is not required for the procedure. The final result is a list of $k$ centers. The clusters are found by assigning each point to its closest center. This configuration can serve as a starting solution to improvement algorithms such as k-means, \citep{L57}, location-allocation \citep[{\color{black}ALT,} ][]{Co63,Co64}, {\color{black}\citep[IALT (Improved ALT) ][]{BD12}}, or the improvement algorithm detailed in Section \ref{sec42}.
	
	The complexity can be reduced by storing for each cluster the minimum increase in the value of the objective function when combined with other clusters. Also, for each cluster we store a cluster number that yields the minimum increase. When two clusters are combined, the minimum increase for all clusters related to the two clusters is recalculated. Also, the increase in the value of the objective function when combined with the newly-formed cluster is checked for all other clusters and if it is smaller than the minimum increase saved for a particular cluster, it replaces the minimum increase for that cluster and the cluster it is combined with. The number of related clusters is expected to be small and if this number does not depend on $n$ and thus does not affect the complexity, the complexity is reduced to  $O(n^2d)$.
	
	The efficient merging algorithm used in the computational experiments is:

	\begin{enumerate}
		\item $n$ clusters are defined, each containing one demand point. Set $m_i=1$
		for  $i=1,\ldots,n$.
		\item For each cluster, in order, calculate by Theorem \ref{Th3} the minimum increase in the value of the objective function $\Delta_i$ and the cluster $j(i)\ne i$ for which this minimum is obtained. If $\Delta_i=0$ is encountered for cluster $j$ (which must be $j>i$), combine clusters $i$ and $j$ creating a revised cluster $i$. Note that the center of the revised cluster is unchanged since $\Delta_i=0$. Replace cluster $j$ with the last cluster in the list and reduce the list of clusters by 1. Continue the procedure  with the evaluation of $\Delta_i$ (from cluster \#1 upward) for the combined cluster $i$.
		\item Repeat the following until the number of clusters is reduced to $k$.
		\begin{enumerate}
			\narrow
			\item Find $i$ for which $\Delta_i$ is minimized (applying GRASP). 
			\item Combine clusters $i$ and $j(i)$, find its center by {\color{black}equation} (\ref{case3}) and replace $m_i$ by $m_i+m_j$. 
			\item Replace cluster $i$ by the combined cluster and remove cluster $j(i)$. The number of clusters is reduced by one.
			\item For each cluster $k\ne i$:
			\begin{enumerate}
				\item Calculate the increase in the value of the objective function by combining cluster $k$ and the combined cluster $i$. If the increase is less than $\Delta_k$, update $\Delta_k$, set $j(k)=i$ and proceed to the next $k$.
				\item If $j(k)=i \mbox{ or }j(i)$, recalculate $\Delta_k$ and $j(k)$.	
				
			\end{enumerate}
			\item Find $\Delta_i$ and $j(i)$  for the combined cluster.
			
		\end{enumerate}
	\end{enumerate}
	
	\subsubsection{Testing the factor $\alpha$ in GRASP}
	
	In the computational experiments we tested ten problems using $\alpha=1.1, 1.5, 2$. Run time increases with larger values of $\alpha$ because more iterations are required by the improvement algorithm. In nine of the ten problems $\alpha=1.5$ and $\alpha=2$ provided comparable results and $\alpha=1.1$ provided somewhat poorer results. However, {\color{black}one data set (kegg)}, exhibited a different behavior. The comparison results are depicted in Table \ref{53}. See also Table \ref{1000} below. For this problem $\alpha=2$ performed the worst.
	
	\begin{table}[ht!]
		\begin{center}
			\caption{\label{53}Results for the {\color{black}kegg data set} by various approaches (100 runs)}
			\medskip
			
			\setlength{\tabcolsep}{3.5pt}
			\begin{tabular}{|c|c||c|c|c||c|c|c||c|c|c||c|c|c|}
				\hline
					$k$&Best&&&&\multicolumn{3}{|c||}{Merging $\alpha=1.1$}&\multicolumn{3}{|c||}{Merging $\alpha=1.5$}&\multicolumn{3}{|c|}{Merging $\alpha=2.0$}\\
				\cline{6-14}			
				&Known$\dagger$&(1)$^*$&(2)$^*$&(3)$^*$&$^*$&+&$\ddagger$&$^*$&+&$\ddagger$&$^*$&+&$\ddagger$\\
				\hline
2&	1.13853E+09&	0.00&	0.00&	0.00&	0.00&	75&	25.1&	0.00&	24&	27.5&	0.00&	16&	30.7\\
5&	1.88367E+08&	0.00&	0.00&	0.00&	0.00&	100&	26.9&	0.00&	100&	30.5&	0.00&	96&	35.4\\
10&	6.05127E+07&	4.96&	0.00&	36.81&	0.00&	22&	33.9&	0.00&	16&	41.0&	0.00&	5&	50.9\\
15&	3.49362E+07&	0.53&	4.00&	98.23&	0.08&	1&	45.3&	0.00&	1&	56.6&	1.20&	1&	64.0\\
20&	2.47108E+07&	1.12&	13.74&	136.71&	0.03&	1&	45.8&	0.01&	1&	56.8&	1.39&	1&	66.0\\
25&	1.90899E+07&	1.27&	15.48&	190.95&	0.31&	1&	58.4&	0.53&	1&	82.6&	1.79&	1&	129.1\\
				\hline							
				\multicolumn{14}{l}{$\dagger$ Best known solution \citep{GV19}.}\\
					\multicolumn{14}{l}{$^*$ Percent above best known solution {\color{black} (relative error)}.}\\
					\multicolumn{14}{l}{(1) Best solution of \citep{BODX15}.}\\
					\multicolumn{14}{l}{(2) Best of the three procedures available in R using the ``++" starting solution.}\\
					\multicolumn{14}{l}{(3) Best of the three procedures available in R from a random starting solution.}\\
				\multicolumn{14}{l}{+ Number of times (out of 100) that the best solution found by the particular $\alpha$ was observed.}\\
								\multicolumn{14}{l}{$\ddagger$ Time in seconds for one run.}\\
			\end{tabular}
		\end{center}
	\end{table}
	
	\subsection{Construction \citep{KGD19}}
	
\citet{KGD19} designed a construction algorithm to solve a clustering problem with a different objective. Rather than minimizing the sum of the squares of distances of all the points from their closest center, their objective is to minimize the sum of squares of distances between all pairs of points belonging to the same cluster. This is an appropriate objective if the points are, for example, sport teams and the clusters are divisions. All pairs of teams in a division play against the other teams in the same division, so the total travel distance by teams is minimized.

	The first two phases generate a starting solution and the third phase is an improvement algorithm that can be replaced by any improvement algorithm. It can also serve as an improvement algorithm for other starting solutions.

For the ``GRASP" approach we propose to find the best move and the second best move. The best move is selected with probability $\frac23$ and the second best with probability $\frac13$. This random selection rule is indicated in the algorithms below as ``applying  GRASP". If no random component is desired, the ``applying GRASP" operation should be ignored.  Other GRASP approaches can be used as well.

	\begin{description}
		
		\item[Phase 1 (selecting one point for each cluster):]
		~
		\begin{itemize}
			\item Randomly select two points. One is assigned to cluster \#1 and one to cluster \#2. (Note that we tested the selection of the farthest or second farthest two points and obtained inferior results.)
			\item Select for the next cluster the point for which the minimum distance to the already selected points is the largest or second-largest (applying GRASP).
			\item Continue until $k$ points define $k$ cluster centers. 
		\end{itemize}

		\item[Phase 2 (adding additional points to clusters):]
		~
		\begin{itemize}
			\item Check all unassigned points to be added to each of the clusters. Select the point that adding it to one of the clusters increases the objective function the least or second-least (applying GRASP) and add this point to the particular cluster.
			\item Keep adding points to clusters until all the points are assigned to a cluster.
		\end{itemize}
		
		\item[Phase 3 (a descent algorithm):]
		~
		\begin{enumerate}
			\item \label{st31} Evaluate all combinations of moving a point from their assigned cluster to another one.
			\item If an improvement is found, perform the best move and go to Step \ref{st31}.
			\item If no improvement is found, stop.
		\end{enumerate}

	\end{description}

Phase 3 is similar to the improvement procedure in {\color{black} \citet{HW79}}.	
The complexity of Phase 1 is $O(nk^2d)$. Phase 2 is repeated $n-k$ times and each time is of complexity $O(nkd)$ once we save for each cluster $j$, its center and its individual objective. The complexity of the first two phases (finding the starting solution) is $O(n^2kd)$ because evaluating the value of the objective function is of complexity $O(d)$ by Theorem~\ref{Th4}. Phase 3 is an improvement algorithm and can be replaced by other improvement algorithms, such as k-means. The complexity of each iteration is $O(nkd)$ by Theorem \ref{Th4}.  In our computational experiments we used the procedure proposed in Section \ref{sec42} which includes Phase~3. 

There are similarities between Phase 1 of the  construction algorithm and the k-means++ algorithm \citep{David2007}. While in Phase 1 we select the added cluster among the farthest and second farthest points, k-means++ selects such a point, among all points,  whose probability is proportional to the squared distance to the closest cluster. k-means++ then applies an improvement algorithm while our proposed construction algorithm has two additional phases.

	\subsection{Separation \citep{BD19}}

	This algorithm finds a starting solution by solving many smaller clustering problems.
	Suppose that $n$ points exist in an area and $k$ cluster centers need to be found. We select $q<k$, for example, $q=\sqrt k$ rounded could work well. We then solve the problem using $q$ clusters by a heuristic or an optimal algorithm. Each of the $q$ centers has a subset allocated to it. We treat these subsets as clusters. It should be noted that for two dimensional problems there are straight lines separating the clusters as in a Voronoi diagram \citep{OKSC00}. For higher dimensions these lines are hyper-planes. This means that the plane is partitioned into polygons (polyhedra in higher dimensions) and a center is located ``near the center" of each polyhedron. All the points inside a polyhedron are closest to the center located at that polyhedron.
	
	We then assign the $k$ centers among the $q$ polyhedra by the procedures described in \citet{BD19}. It is expected that many clusters will get about the same number of points, and hence,  many sub-problems will have roughly $\frac k q$ clusters and $\frac nq$ points.
	Once the allocation is complete, a solution is found and it can serve as a starting solution to a variety of heuristic algorithms.
	
	We applied the construction algorithm for solving the small problems because smaller problems have significantly lower computation time. The merging algorithm requires similar run times for varying values of $k$ and the same $n$. Therefore, the first phase of finding $q$ clusters takes about the same time as solving the complete problem with $k$ centers, and there would be no time saving.
	
	\subsubsection{Selecting $q$}
	
	The complexity of the construction algorithm is $O(n^2kd)$. Assume that the time required for running the algorithm is $\beta n^2kd$ for some constant $\beta$.  We first create $q$ clusters by solving the $q$ clusters problem. The time for this algorithm is $\beta n^2qd$. Then, once $q$ clusters are determined, we solve $k+q-1$ problems ($q$ of them of one cluster which is found in $O(nd)$ time) each having about $\frac nq$ points and up to $\frac kq$ clusters each  for a total time of $k\beta\frac{n^2}{q^2}\frac kq d $. The total time is $\beta n^2qd+\beta\frac{n^2k^2}{q^3}d$. The variable term (dividing by $\beta n^2d$) is $q+\frac{k^2}{q^3}$ whose minimum is obtained for $q=\sqrt[4]{3}\sqrt k\approx 1.3\sqrt{k}$. Since many of the $k+q-1$ problems consist of fewer than $\frac k q$ clusters, this value should be rounded down to $q=\lfloor 1.3\sqrt k\rfloor$. For example, for $k=10$ we should choose $q=4$ and for $k=25$ we should choose $q=6$. Note that this procedure for selecting $q$ aims at getting an efficient running time of the algorithm, and not necessarily the best quality of the solution.

	\section{\label{refine}The Proposed Improvement Algorithm}
	
	The k-means improvement procedure \citep{L57} is the same as the location-allocation improvement algorithm \citep{Co63,Co64}.
	
	\begin{enumerate}
\item Select an initial set of $k$ centers.
\item \label{stp2} Allocate each point to its closest center forming $k$ clusters.
\item If the clusters did not change, stop. Otherwise, find the optimal location for the center of each cluster and go to Step \ref{stp2}

	\end{enumerate}
	
	\subsection{\label{sec41}Comparing Phase 3 of the Construction Algorithm to k-means}
	
 When k-means terminates, Phase 3 of the construction algorithm may find improved solutions. Let $d_i^2(X_j)$ be the squared Euclidean distance between point $i$ and $X_j$ (the center of cluster $j$), and let cluster $j$  have $m_j$ points. The location-allocation algorithm stops when every point is assigned to its closest center. Phase~3 may find a better solution when there exists a move of point $i$ from its closest cluster $j_1$ to another cluster $j_2$ so that the objective function improves. Of course, $d_i^2(X_{j_1})<d_i^2(X_{j_2})$ because the k-means terminated. However, such a move improves the value of the objective function if by Theorems \ref{Th1},\ref{Th2}:
	\begin{equation}\label{ratio}\frac{m_{j_1}}{m_{j_1}-1}d_i^2(X_{j_1})>\frac{m_{j_2}}{m_{j_2}+1}d_i^2(X_{j_2})
	~\rightarrow~d_i^2(X_{j_2})~<~\frac{1+\frac1{m_{j_2}}}{1-\frac1{m_{j_1}}}d_i^2(X_{j_1})
	\end{equation}
	For example, for clusters of sizes $m_1=m_2=5$, the squared distance to the center of the second cluster can be up to 50\% larger. The objective function will improve by such a move if the squared distance to the center of the second cluster is less than 1.5 times the squared distance to the center of the closest cluster.
	
\begin{figure}[ht!]
		\setlength{\unitlength}{1in}
		\centering	\begin{picture}(3,1.2)
		\put(0,0){\circle*{0.1}}
		\put(1,1){\circle*{0.1}}		
		\put(0,1){\circle*{0.1}}		
		\put(1,0){\circle*{0.1}}		
		\put(1.366,0){\circle*{0.1}}		
		\put(1.366,1){\circle*{0.1}}		
		\put(2.366,0){\circle*{0.1}}		
		\put(2.366,1){\circle*{0.1}}
		\put(1,1){\line(1,0){0.366}}
		\put(1.13,1.05){$x$}
		\put(1.366,1.1){$A$}
		\put(1.366,0.1){$B$}
		\put(2.366,1.1){$C$}
		\put(-0.4,0){(0, 0)}						
		
		\end{picture}
		\caption{The squares example. {\color{black}$x$ is the distance between the squares.}}
		\label{two}
	\end{figure}
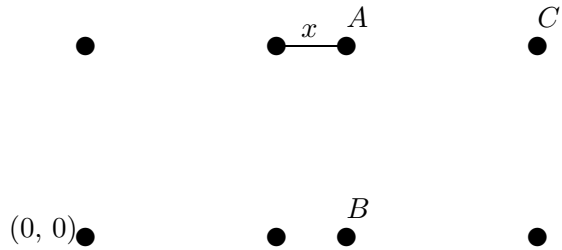

	As an example, consider two squares of side 1 with vertices as the given points. The squares' close sides are $x$ units apart (see Figure \ref{two}). Two clusters are sought. There are two ``natural" cluster centers at the centers of the squares, each cluster with 4 points for an objective of 4. This is one of the possible final solutions of k-means. Each given point is closest to the center of its assigned cluster of 4 points.

	The four vertices closest to the other square, for example {\color{black} vertices} $A$ and $B$, are at squared distance of $\frac14+(\frac12+x)^2=\frac12+x+x^2$ from the farther center. Suppose that in Phase 3 point $A$ is ``moved" to the cluster of the left square. The center of the left cluster is moved to $(0.6+\frac15 x,0.6)$ and the center of the right cluster is moved to $(\frac53 +x ,\frac13 )$ for a total objective of $2.4+\frac45x+\frac45x^2  +\frac43=3\frac8{15}+\frac15(1+2x)^2$. This objective is less than 4 for $x<\sqrt{\frac{7}{12}}-\frac12=0.2638$. Point $A$ is at squared distance of $\frac12$ from the right square center, and is at squared distance $\frac56$ from the left square center for the largest $x$. It is greater by a factor of $\frac53$ confirming the factor obtained by equation (\ref{ratio})  substituting $m_1=m_2=4$. If we move point $B$ to the left cluster to have clusters of 6 and 2 points, the ratio of the squared distances by equation (\ref{ratio}) with $m_1=3$, $m_2=5$ is  $\frac95$ indicating that for some values of $x$ two clusters of 6 and 2 are even better. Its objective is $\frac{10}{3}+\frac43 (x+x^2)=3+\frac13(1+2x)^2$ which is less than 4 for $x<\frac12(\sqrt3-1)=0.366$. This configuration for the largest $x$ is depicted in Figure \ref{two}. It is better than the 5-3 configuration for $x$ as high as 0.5, practically for all values of $x$ which are of interest. The objective by adding point $C$ to the left cluster for a 7-1 configuration is greater than the 6-2 objective  for any $x\ge 0$ and thus it is never optimal.
	
	Note that there are many cluster partitions that are an inferior  terminal solution to the clustering two square problem. For example, the solution with clusters consisting of the top 4 points and the bottom 4 points has an objective of $4+4x+2x^2>4$, and cannot be improved by k-means.
	
	We ran the eight point problem for $x=0.25$ by the construction algorithm 1000 times. It found the optimal solution of $3\frac34$ in 873 of the 1000 starting solutions. For $x=0.3$ the optimal solution of $3\frac{64}{75}$ was found 856 times out of 1000. For $x=0.35$ the optimal solution of $3\frac{289}{300}$ was found 840 times out of 1000. The algorithm was so fast that we had to run it a million times to get measurable run time. Run time is 0.86 millionth of a second for one run.
	
	This performance is compared in Table \ref{comp} with commonly used procedures detailed in the introduction. The merging algorithm clearly performed best and the construction algorithm was second-best. Hartigan-Wong performed better than the other commonly used procedures. Note that the four commonly used procedures started from the same 1000 starting solutions.
	
	\begin{table}[htp!]
		\begin{center}
			\caption{\label{comp}Comparing results for the two squares problem}
			\setlength{\tabcolsep}{7pt}
			\medskip
			
			\begin{tabular}{|l|c|c|c|}
				\hline
				Method&$x=0.25$&$x=0.30$&$x=0.35$\\
				\cline{2-4}
				&\multicolumn{3}{|c|}{Optimum$\dagger$ found in 1,000 runs}\\
				\hline
				Merging&1,000&1,000&1,000\\
				Construction&873&856&840\\
				Forgy&348&317&347\\
				Hartigan-Wong&794&479&487\\		
				Lloyd&348&317&347\\
				MacQueen&348&317&347\\
				\hline
				\multicolumn{4}{l}{$\dagger$ optimal value is $3+\frac13(1+2x)^2$}
			\end{tabular}
		\end{center}
	\end{table}
	
	It is interesting that the merging algorithm starting solution is the optimal solution to the two square example. In the first two iterations the two pairs of close points will be combined into two clusters. Then, in the next two iterations the two right points and the two left points will be combined into two clusters yielding after four iterations four clusters of two points each forming a rhombus with centers at the midpoints of the four sides of the rectangle enclosing the two squares. In the fifth iteration the top and bottom clusters will be combined forming a cluster of 4 points at the center of the configuration. There are three centers on the central horizontal line. In the sixth iteration the central cluster will be combined with one of the other two forming two clusters of 6 and 2 points which is the optimal solution. 
	
	\subsection{\label{sec42}The Improvement Algorithm Used in the Experiments}
	
One iteration of both the k-means and Phase 3 has the same complexity of $O(nkd)$.	
Since the k-means algorithm may move several points from one cluster to another, it is faster than Phase~3 because fewer iterations are required. However, as is shown in Section \ref{sec41}, Phase 3 may improve terminal results of k-means. We therefore start by applying k-means up to 10 iterations and then apply Phase 3 until it terminates.

	\begin{table}[htp!]
		\begin{center}
			\caption{\label{prob}Test {\color{black}Data Sets}}
			\setlength{\tabcolsep}{7pt}
			\medskip
			
			\begin{tabular}{|l||c|c|l|}
				\hline
			{\color{black}Name}&	$n$&$d$&Source\\
				\hline
				\multicolumn{4}{|c|}{Small Problems}\\
				\hline
				{\color{black}ruspini75}&75&2&\cite{PABM19}\\
				{\color{black}fisher}&150&4&\cite{BODX15,PABM19}\\
				{\color{black}gr202}&202&2&\cite{PABM19}\\
				{\color{black}gr666}&666&2&\cite{PABM19}\\
				\hline
				\multicolumn{4}{|c|}{Medium Size Problems}\\
				\hline
				{\color{black}tsplib1060}&1,060&2&\cite{BODX15,PABM19}\\
				{\color{black}tsplib3038}&3,038&2&\cite{BODX15,PABM19}\\
			{\color{black}pendigit}&10,992&16&\cite{BODX15}\\
				\hline
				\multicolumn{4}{|c|}{Large Problems}\\
				\hline
				{\color{black}letter}&20,000&16&\cite{BODX15}\\
			{\color{black}kegg}&53,413&20&\cite{BODX15}\\
				{\color{black}pla85900}&85,900&2&\cite{BODX15,PABM19}$\dagger$\\
				\hline
\multicolumn{4}{l}{\small $\dagger$ Results in \citep{PABM19} for this problem are wrong.}\\\multicolumn{4}{l}{~~\small Only a partial data set is accidentally used.}				
			\end{tabular}
		\end{center}
	\end{table}

	\begin{table}[ht!]
		\begin{center}
			\caption{\label{p5}Results for Problems with $k\le 5$ by the Merging and Construction Methods}
			\medskip
			\setlength{\tabcolsep}{5pt}	
			\begin{tabular}{|c|c|c|c|c||c|c|c|c|c|}
				\hline
				Data&$k$&Best$\dagger$&\multicolumn{2}{|c||}{Time (sec.)$\ddagger$}&Data&$k$&Best$\dagger$&\multicolumn{2}{|c|}{Time (sec.)$\ddagger$}\\
\cline{4-5}
\cline{9-10}
				Set&&Known&(1)&(2)&Set&&Known&(1)&(2)\\
				\hline
{\color{black}ruspini75}&	2&	8.93378E+04&	0.0001&	0.0002&	{\color{black}gr666}&	4&	6.13995E+05&	0.0067&	0.0108\\
{\color{black}ruspini75}&	3&	5.10634E+04&	0.0001&	0.0002&	{\color{black}gr666}&	5&	4.85088E+05&	0.0086&	0.0108\\
{\color{black}ruspini75}&	4&	1.28810E+04&	0.0001&	0.0002&	{\color{black}tsplib1060}&	2&	9.83195E+09&	0.0093&	0.0310\\
{\color{black}ruspini75}&	5&	1.01267E+04&	0.0001&	0.0002&	{\color{black}tsplib1060}&	5&	3.79100E+09&	0.025&	0.032\\
{\color{black}fisher}&	2&	1.52348E+02&	0.0002&	0.0009&	{\color{black}tsplib3038}&	2&	3.16880E+09&	0.079&	0.243\\
{\color{black}fisher}&	3&	7.88514E+01&	0.0003&	0.0009&	{\color{black}tsplib3038}&	5&	1.19820E+09&	0.193&	0.248\\
{\color{black}fisher}&	4&	5.72285E+01&	0.0004&	0.0009&	{\color{black}pendigit}&	2&	1.28119E+08&	1.51&	6.75\\
{\color{black}fisher}&	5&	4.64462E+01&	0.0005&	0.0009&	{\color{black}pendigit}&	5&	7.53040E+07&	3.90&	6.59\\
{\color{black}gr202}&	2&	2.34374E+04&	0.0004&	0.0013&	{\color{black}letter}&	2&	1.38189E+06&	7.38&	19.37\\
{\color{black}gr202}&	3&	1.53274E+04&	0.0005&	0.0013&	{\color{black}letter}&	5&	1.07712E+06&	23.47&	21.71\\
{\color{black}gr202}&	4&	1.14556E+04&	0.0008&	0.0013&	{\color{black}kegg}&	2&	1.13853E+09&	205.30&	27.47\\
{\color{black}gr202}&	5&	8.89490E+03&	0.0009&	0.0013&	{\color{black}kegg}&	5&	1.88367E+08&	300.09&	30.52\\
{\color{black}gr666}&	2&	1.75401E+06&	0.0035&	0.0107&	{\color{black}pla85900}&	2&	3.74908E+15&	84.51&	193.81\\
{\color{black}gr666}&	3&	7.72707E+05&	0.0052&	0.0107&	{\color{black}pla85900}&	5&	1.33972E+15&	168.16&	200.15\\

				\hline							
				\multicolumn{10}{l}{$\dagger$ Best known solution was found for all instances.}\\
				\multicolumn{10}{l}{$\ddagger$ Time in seconds for one run.}\\
				\multicolumn{10}{l}{(1) Run time by the construction method.}\\
				\multicolumn{10}{l}{(2) Run time by the merging method using $\alpha=1.5$.}\\
			\end{tabular}
		\end{center}
	\end{table}

		\begin{table}[ht!]
		\begin{center}
			\caption{\label{sm}Results for small Problems with $k> 5$ (1000 runs)}
			\medskip
			\small
			\setlength{\tabcolsep}{4pt}
			\begin{tabular}{|c|c||c||c|r|c|r|c|c|c|c|}
				\hline
				Data&$k$&Best&\multicolumn{2}{|c|}{$\alpha=1.5\S$}&\multicolumn{2}{|c|}{$\alpha=2.0\S$}&\multicolumn{2}{|c|}{Construction}&\multicolumn{2}{|c|}{Separation}\\
				\cline{4-11}			
				Set&&Known&$^*$&Time$\ddagger$&$^*$&Time$\ddagger$&$^*$&Time$\ddagger$&$^*$&Time$\ddagger$\\
				\hline
{\color{black}ruspini75}&	6&	8.57541E+03&	0.00&	0.0002&	0.00&	0.0003&	0.00&	0.0001&	0.00&	0.0002\\
{\color{black}ruspini75}&	7&	7.12620E+03&	0.00&	0.0003&	0.00&	0.0003&	0.00&	0.0001&	0.00&	0.0002\\
{\color{black}ruspini75}&	8&	6.14964E+03&	0.00&	0.0002&	0.00&	0.0003&	0.00&	0.0002&	0.00&	0.0002\\
{\color{black}ruspini75}&	9&	5.18165E+03&	0.00&	0.0002&	0.00&	0.0003&	0.00&	0.0002&	0.00&	0.0002\\
{\color{black}ruspini75}&	10&	4.44628E+03&	0.00&	0.0003&	0.00&	0.0003&	0.00&	0.0002&	0.00&	0.0002\\
{\color{black}fisher}&	6&	3.90400E+01&	0.00&	0.0009&	0.00&	0.0012&	0.00&	0.0006&	0.00&	0.0007\\
{\color{black}fisher}&	7&	3.42982E+01&	0.00&	0.0009&	0.00&	0.0012&	0.00&	0.0006&	0.00&	0.0007\\
{\color{black}fisher}&	8&	2.99889E+01&	0.00&	0.0010&	0.00&	0.0012&	0.00&	0.0007&	0.00&	0.0007\\
{\color{black}fisher}&	9&	2.77861E+01&	0.00&	0.0010&	0.00&	0.0013&	0.00&	0.0008&	0.00&	0.0009\\
{\color{black}fisher}&	10&	2.58340E+01&	0.00&	0.0010&	0.00&	0.0013&	0.06&	0.0009&	0.00&	0.0008\\
{\color{black}gr202}&	6&	6.76488E+03&	0.00&	0.0013&	0.00&	0.0016&	0.00&	0.0011&	0.00&	0.0014\\
{\color{black}gr202}&	7&	5.81757E+03&	0.00&	0.0014&	0.00&	0.0016&	0.00&	0.0012&	0.00&	0.0015\\
{\color{black}gr202}&	8&	5.00610E+03&	0.00&	0.0014&	0.00&	0.0017&	0.03&	0.0013&	0.00&	0.0014\\
{\color{black}gr202}&	9&	4.37619E+03&	0.00&	0.0014&	0.00&	0.0017&	0.77&	0.0014&	0.00&	0.0019\\
{\color{black}gr202}&	10&	3.79449E+03&	0.00&	0.0015&	0.00&	0.0018&	0.00&	0.0016&	0.00&	0.0016\\
{\color{black}gr666}&	6&	3.82676E+05&	0.00&	0.0108&	0.00&	0.0133&	0.00&	0.0106&	0.00&	0.0126\\
{\color{black}gr666}&	7&	3.23283E+05&	0.00&	0.0109&	0.00&	0.0133&	0.00&	0.0124&	0.00&	0.0125\\
{\color{black}gr666}&	8&	2.85925E+05&	0.00&	0.0110&	0.00&	0.0134&	0.00&	0.0143&	0.00&	0.0123\\
{\color{black}gr666}&	9&	2.50989E+05&	0.00&	0.0110&	0.00&	0.0137&	0.13&	0.0163&	0.00&	0.0163\\
{\color{black}gr666}&	10&	2.24183E+05&	0.00&	0.0113&	0.00&	0.0140&	0.20&	0.0180&	0.00&	0.0131\\
				\hline							
				\multicolumn{11}{l}{$\S$ Variant of the merging procedure}\\
				\multicolumn{11}{l}{$\ddagger$ Time in seconds for one run.}\\
				\multicolumn{11}{l}{$^*$ Percent above best known solution {\color{black} (relative error)}.}
				
			\end{tabular}
		\end{center}
	\end{table}

	\begin{table}[ht!]
		\begin{center}
			\caption{\label{med}Results for Medium Problems with $k> 5$ (1000 runs)}
			\medskip
			\small
			\setlength{\tabcolsep}{4pt}
			\begin{tabular}{|c|c||c||c|c|r|c|r|c|c|c|c|}
				\hline
			Data&$k$&Best$\dagger$&Prior&\multicolumn{2}{|c|}{$\alpha=1.5\S$}&\multicolumn{2}{|c|}{$\alpha=2.0\S$}&\multicolumn{2}{|c|}{Construction}&\multicolumn{2}{|c|}{Separation}\\
				\cline{5-12}			
				Set&&Known&BK$^*$&$^*$&Time$\ddagger$&$^*$&Time$\ddagger$&$^*$&Time$\ddagger$&$^*$&Time$\ddagger$\\
				\hline
{\color{black}tsplib1060}&	10&	1.75484E+09&	0.00&	0.00&	0.03&	0.00&	0.04&	0.00&	0.05&	0.00&	0.03\\
{\color{black}tsplib1060}&	15&	1.12114E+09&	0.00&	0.00&	0.04&	0.01&	0.04&	0.02&	0.08&	0.02&	0.04\\
{\color{black}tsplib1060}&	20&	7.91790E+08&	0.00&	0.00&	0.04&	0.01&	0.04&	0.06&	0.10&	0.00&	0.05\\
{\color{black}tsplib1060}&	25&	{\bf 6.06607E+08} &	0.02&	0.03&	0.04&	0.02&	0.04&	0.46&	0.12&	0.10&	0.06\\
{\color{black}tsplib3038}&	10&	5.60251E+08&	0.00&	0.00&	0.27&	0.00&	0.32&	0.00&	0.41&	0.00&	0.24\\
{\color{black}tsplib3038}&	15&	3.56041E+08&	0.00&	0.00&	0.30&	0.00&	0.37&	0.00&	0.67&	0.00&	0.33\\
{\color{black}tsplib3038}&	20&	2.66812E+08&	0.00&	0.01&	0.34&	0.03&	0.41&	0.01&	0.93&	0.00&	0.38\\
{\color{black}tsplib3038}&	25&	{\bf 2.14475E+08} &	0.01&	0.02&	0.37&	0.02&	0.44&	0.11&	1.15&	0.00&	0.44\\
{\color{black}pendigit}&	10&	4.93015E+07&	0.00&	0.00&	6.97&	0.00&	10.31&	0.00&	8.03&	0.00&	5.48\\
{\color{black}pendigit}&	15&	3.90675E+07&	0.00&	0.00&	7.00&	0.00&	10.37&	0.00&	10.49&	0.00&	6.96\\
{\color{black}pendigit}&	20&	3.40194E+07&	0.00&	0.00&	7.26&	0.00&	10.77&	0.14&	14.66&	0.00&	8.30\\
{\color{black}pendigit}&	25&	{\bf 2.99865E+07} &	0.17&	0.00&	7.38&	0.00&	10.95&	0.75&	18.74&	0.69&	8.93\\
				\hline							
				\multicolumn{12}{l}{$\dagger$ Best known solution including the results in this paper. New ones displayed in boldface.}\\
				\multicolumn{12}{l}{$\S$ Variant of the merging procedure}\\
				\multicolumn{12}{l}{$\ddagger$ Time in seconds for one run.}\\
				\multicolumn{12}{l}{$^*$ Percent above best known solution {\color{black} (relative error)}.}
				
			\end{tabular}
		\end{center}
	\end{table}
		
\begin{table}[ht!]
	\begin{center}
		\caption{\label{large}Results for Large Problems with $k> 5$ (100 runs)}
		\medskip
		\small
		\setlength{\tabcolsep}{4pt}
		\begin{tabular}{|c|c||c||c|c|r|c|r|c|c|c|c|}
			\hline
		Data&$k$&Best$\dagger$&Prior&\multicolumn{2}{|c|}{$\alpha=1.5\S$}&\multicolumn{2}{|c|}{$\alpha=2.0\S$}&\multicolumn{2}{|c|}{Construction}&\multicolumn{2}{|c|}{Separation}\\
			\cline{5-12}			
			Set&&Known&BK$^*$&$^*$&Time$\ddagger$&$^*$&Time$\ddagger$&$^*$&Time$\ddagger$&$^*$&Time$\ddagger$\\
			\hline
{\color{black}letter}&	10&	8.57503E+05&	0.00&	0.00&	22.9&	0.00&	34.4&	0.00&	37.1&	0.00&	29.0\\
{\color{black}letter}&	15&	{\bf 7.43923E+05} &	0.09&	0.00&	25.1&	0.00&	38.1&	0.00&	57.2&	0.00&	36.0\\
{\color{black}letter}&	20&	6.72593E+05&	0.00&	0.16&	30.6&	0.00&	44.4&	0.20&	78.1&	0.01&	41.7\\
{\color{black}letter}&	25&	{\bf 6.19572E+05} &	0.53&	0.00&	36.1&	0.00&	49.0&	0.50&	96.7&	0.13&	42.2\\
{\color{black}kegg}&	10&	6.05127E+07&	0.00&	0.00&	41.0&	0.00&	50.9&	4.96&	401.2&	0.00&	673.3\\
{\color{black}kegg}&	15&	3.49362E+07&	0.00&	0.00&	56.6&	1.20&	64.0&	11.79&	513.5&	10.34&	824.2\\
{\color{black}kegg}&	20&	2.47108E+07&	0.00&	0.01&	56.8&	1.39&	66.0&	13.09&	697.8&	12.15&	984.1\\
{\color{black}kegg}&	25&	1.90899E+07&	0.00&	0.53&	82.6&	1.79&	129.1&	29.53&	784.0&	15.60&	878.0\\
{\color{black}pla85900}&	10&	6.82941E+14&	0.00&	0.00&	251.6&	0.00&	288.7&	0.00&	379.0&	0.00&	202.2\\
{\color{black}pla85900}&	15&	4.60294E+14&	0.00&	0.00&	325.4&	0.00&	385.6&	0.00&	680.4&	0.00&	264.1\\
{\color{black}pla85900}&	20&	3.49811E+14&	0.00&	0.01&	437.4&	0.00&	509.6&	0.00&	984.5&	0.00&	336.9\\
{\color{black}pla85900}&	25&	2.82217E+14&	0.00&	0.00&	490.4&	0.02&	547.2&	0.00&	1180.2&	0.00&	377.8\\
			\hline							
			\multicolumn{12}{l}{$\dagger$ Best known solution including the results in this paper. New ones displayed in boldface.}\\
			\multicolumn{12}{l}{$\S$ Variant of the merging procedure}\\
			\multicolumn{12}{l}{$\ddagger$ Time in seconds for one run.}\\
			\multicolumn{12}{l}{$^*$ Percent above best known solution {\color{black} (relative error)}.}
			
		\end{tabular}
	\end{center}
\end{table}

	\begin{table}[ht!]
	\begin{center}
		\caption{\label{Rsml}Percent above best known solution by the procedures in R for small problems}
		\medskip
		
		\setlength{\tabcolsep}{4pt}
		\begin{tabular}{|c|c||c||c|c||c|c||c|c|}
			\hline
		Data&$k$&Best&\multicolumn{2}{|c||}{H-W}&\multicolumn{2}{|c||}{Lloyd}&\multicolumn{2}{|c|}{MacQ}\\
			\cline{4-9}			
			Set&&Known$\dagger$&Rand&++&Rand&++&Rand&++\\
			\hline
{\color{black}ruspini75}&	6&	8.57541E+03&	0.00&	0.00&	0.00&	0.00&	0.00&	0.00\\
{\color{black}ruspini75}&	7&	7.12620E+03&	0.00&	0.00&	0.00&	1.34&	0.00&	1.34\\
{\color{black}ruspini75}&	8&	6.14964E+03&	0.00&	0.00&	0.00&	1.61&	0.00&	1.61\\
{\color{black}ruspini75}&	9&	5.18165E+03&	0.00&	2.17&	0.00&	3.78&	0.00&	5.56\\
{\color{black}ruspini75}&	10&	4.44628E+03&	0.00&	0.38&	1.25&	13.99&	0.48&	13.99\\
{\color{black}fisher}&	6&	3.90400E+01&	0.00&	0.00&	0.00&	0.00&	0.00&	0.00\\
{\color{black}fisher}&	7&	3.42982E+01&	0.00&	0.00&	0.00&	0.00&	0.00&	0.00\\
{\color{black}fisher}&	8&	2.99889E+01&	0.00&	0.00&	0.01&	0.01&	0.00&	0.01\\
{\color{black}fisher}&	9&	2.77861E+01&	0.00&	0.00&	0.01&	0.39&	0.01&	0.39\\
{\color{black}fisher}&	10&	2.58341E+01&	0.00&	0.06&	0.16&	0.78&	0.16&	0.78\\
{\color{black}gr202}&	6&	6.76488E+03&	0.00&	0.00&	0.00&	0.00&	0.00&	0.00\\
{\color{black}gr202}&	7&	5.81757E+03&	0.00&	0.00&	0.01&	0.02&	0.02&	0.12\\
{\color{black}gr202}&	8&	5.00610E+03&	0.00&	0.00&	0.03&	0.07&	0.03&	0.09\\
{\color{black}gr202}&	9&	4.37619E+03&	0.00&	0.00&	0.00&	0.00&	0.00&	0.00\\
{\color{black}gr202}&	10&	3.79449E+03&	0.00&	0.00&	0.00&	0.58&	0.29&	2.32\\
{\color{black}gr666}&	6&	3.82677E+05&	0.00&	0.00&	0.00&	0.00&	0.00&	0.00\\
{\color{black}gr666}&	7&	3.23284E+05&	0.00&	0.00&	0.00&	0.06&	0.00&	0.00\\
{\color{black}gr666}&	8&	2.85925E+05&	0.00&	0.00&	0.00&	0.00&	0.00&	0.00\\
{\color{black}gr666}&	9&	2.50989E+05&	0.00&	0.00&	0.00&	0.03&	0.00&	0.04\\
{\color{black}gr666}&	10&	2.24184E+05&	0.00&	0.41&	0.00&	0.14&	0.00&	0.41\\
			\hline							
			\multicolumn{9}{l}{$\dagger$ Best known solution including the results in this paper.}\\
		\end{tabular}
	\end{center}
\end{table}

\begin{table}[ht!]
	\begin{center}
		\caption{\label{Rlrg}Percent above best known solution by the procedures in R for medium and large problems}
		\medskip
		
		\setlength{\tabcolsep}{4pt}
		\begin{tabular}{|c|c||c||c|c||c|c||c|c|}
			\hline
			Data&$k$&Best&\multicolumn{2}{|c||}{H-W}&\multicolumn{2}{|c||}{Lloyd}&\multicolumn{2}{|c|}{MacQ}\\
			\cline{4-9}			
			Set&&Known$\dagger$&Rand&++&Rand&++&Rand&++\\
			\hline
{\color{black}tsplib1060}&	10&	1.75484E+09&	0.00&	0.00&	0.00&	0.24&	0.00&	0.25\\
{\color{black}tsplib1060}&	15&	1.12114E+09&	0.00&	4.59&	0.07&	6.86&	0.04&	6.90\\
{\color{black}tsplib1060}&	20&	7.91790E+08&	0.05&	7.06&	0.05&	13.44&	0.21&	12.80\\
{\color{black}tsplib1060}&	25&	6.06607E+08&	0.06&	16.47&	0.37&	23.66&	0.48&	17.90\\
{\color{black}tsplib3038}&	10&	5.60251E+08&	0.00&	0.00&	0.00&	0.00&	0.00&	0.00\\
{\color{black}tsplib3038}&	15&	3.56041E+08&	0.00&	0.01&	0.00&	0.07&	0.00&	0.10\\
{\color{black}tsplib3038}&	20&	2.66812E+08&	0.01&	0.33&	0.03&	0.25&	0.05&	0.49\\
{\color{black}tsplib3038}&	25&	2.14475E+08&	0.04&	1.15&	0.06&	2.15&	0.05&	1.84\\
{\color{black}pendigit}&	10&	4.93015E+07&	0.00&	0.00&	0.00&	0.00&	0.00&	0.00\\
{\color{black}pendigit}&	15&	3.90675E+07&	0.00&	0.00&	0.00&	0.00&	0.00&	0.00\\
{\color{black}pendigit}&	20&	3.40194E+07&	0.09&	0.02&	0.09&	0.03&	0.01&	0.03\\
{\color{black}pendigit}&	25&	2.99865E+07&	0.17&	0.17&	0.05&	0.18&	0.17&	0.18\\
{\color{black}letter}&	10&	8.57503E+05&	0.00&	0.00&	0.00&	0.00&	0.00&	0.00\\
{\color{black}letter}&	15&	7.43923E+05&	0.00&	0.00&	0.00&	0.00&	0.00&	0.00\\
{\color{black}letter}&	20&	6.72593E+05&	0.00&	0.00&	0.00&	0.01&	0.00&	0.01\\
{\color{black}letter}&	25&	6.19572E+05&	0.00&	0.00&	0.10&	0.00&	0.00&	0.01\\
{\color{black}kegg}&	10&	6.05127E+07&	36.81&	0.00&	36.81&	0.00&	36.81&	0.00\\
{\color{black}kegg}&	15&	3.49362E+07&	98.23&	4.41&	98.23&	7.27&	98.23&	4.00\\
{\color{black}kegg}&	20&	2.47108E+07&	136.71&	13.74&	161.55&	19.13&	161.55&	16.17\\
{\color{black}kegg}&	25&	1.90899E+07&	190.95&	15.48&	208.02&	17.68&	208.02&	35.56\\
{\color{black}pla85900}&	10&	6.82941E+14&	0.00&	0.00&	0.00&	0.00&	0.00&	0.00\\
{\color{black}pla85900}&	15&	4.60294E+14&	0.00&	0.00&	0.00&	0.00&	0.00&	0.00\\
{\color{black}pla85900}&	20&	3.49810E+14&	0.00&	0.02&	0.00&	0.00&	0.00&	0.00\\
{\color{black}pla85900}&	25&	2.82215E+14&	0.00&	0.14&	0.00&	0.00&	0.00&	0.00\\
			\hline							
			\multicolumn{9}{l}{$\dagger$ Best known solution including the results in this paper.}\\
		\end{tabular}
	\end{center}
\end{table}

	\section{\label{sec4}Computational Experiments}
	
	We experimented with ten data sets and various numbers of clusters for each. Four problems are classified as small problems, three are medium size and three are large. The problems are listed in Table \ref{prob}. The construction, separation, and merging  algorithms were 
	coded in Fortran using double precision arithmetic and were compiled by an Intel 11.1 Fortran Compiler  using one thread with no parallel processing. They were run on a desktop with the Intel i7-6700 3.4GHz CPU processor and 16GB RAM. Small and medium size problems were run 1000 times each in a multi-start approach. Large problems were run 100 times each.
	
{\color{black}The ``stats" package included in the base installation of ``R" version 3.5.3 contains the k-means function\footnote{{\color{black}The source code of the k-means function is available at\\ https://github.com/SurajGupta/r-source/blob/master/src/library/stats/R/kmeans.R.}}, which implements the three widely-used clustering algorithms described in Section \ref{widely}: Lloyd, MacQueen (MacQ), and Hartigan-Wong (H-W).} 
They were run on a virtualized Windows environment with 16 vCPUs and 128GB of vRAM. No parallel processing was used in R. All algorithms (and starting solutions) were run as a single thread.
	The physical server used was a two PowerEdge R720 Intel E5-2650 CPUs (8 cores each) with 128 GB RAM using shared storage on MD3620i via 10GB interfaces.
	
	Thirty-six instances were tested for small size problems for $k=2,3,\ldots,10$ clusters. Thirty-six instances were tested for medium and large size problems for $k=2,5,10,15,20,25$ clusters, for a total of 72 instances. The best known (for small problems, the optimal) solutions were obtained for the 28 instances of $2\le k\le 5$. Such problems are too small for the separation algorithm so they were tested by the merging and the construction algorithm as well as R's implementations of Lloyd, Hartigan-Wong (H-W), and MacQueen (MacQ) heuristics starting at random solutions, and their "++" starting solution procedure. Run times in seconds for the merging and construction algorithms are depicted in Table \ref{p5}.
	
	We then tested the instances for $k>5$ clusters by the merging, construction, and separation algorithms and R implementations of Lloyd, Hartigan-Wong, and MacQueen heuristics starting at random starting solutions and the ``++" implementation. In Table \ref{sm} the results for small size problems, for which the optimal solution is known \citep{A09}, are reported for the merging, construction, and separation algorithms.
	
	In Table \ref{med} the results for the three medium size problems are reported.  Each instance was solved by two variants of the merging algorithm, the construction and separation algorithms, 1000 times each, and the  best result reported. New best known solutions were found for three of the twelve instances.

	In Table \ref{large} the results for the three large problems are reported. Each instance was solved by the merging, construction and separation algorithms, 100 times, and the  best result reported. New best known solutions were found for two of the twelve instances.
	
	 In Tables \ref{Rsml}, \ref{Rlrg} we report the results by the three R algorithms as well as the special starting solution "++" \citep{David2007} which requires an extra step of preparation of starting solutions. Algorithms were repeated from 1000 starting solutions.
	
	While nine of the ten problems exhibit similar behavior, one large {\color{black} data set (kegg)} exhibited unique behavior. It turns out to be a challenging problem. The merging procedure performed far better than all other approaches. For example, the $k=25$ instance found a solution 190\% worse by the best R approach (see Table \ref{Rlrg}). When the special starting solution ``++" \citep{David2007} was implemented in R, it was ``only'' 15\% higher. The construction algorithm was 29\% higher, the separation was 15\% higher, {\color{black} and the merging procedure was 0.31\% higher (Table \ref{53})}. Also, we tested in the merging procedure $\alpha=1.1, 1.5, 2.0$. In the other nine problems $\alpha=1.5$ and $\alpha=2$ provided comparable results while $\alpha=1.1$ provided somewhat poorer results and thus are not reported. For the {\color{black} data set (kegg)} instances, $\alpha=2$ performed the poorest while $\alpha=1.1$  performed best.

	As a final experiment we solved the {\color{black} data set (kegg)} by the three variants of the merging procedure 1000 times  in order to possibly get further improved best known solutions. The results are depicted in Table \ref{1000}. An improved solution was found for $k=20$.

	\begin{table}[ht!]
	\begin{center}
		\caption{\label{1000}Results for the {\color{black} kegg data set} by merging variants (1000 runs)}
		\medskip
		
		\setlength{\tabcolsep}{4pt}
		\begin{tabular}{|c|c||c|c|c||c|c|c||c|c|c|}
			\hline
				$k$&Best&\multicolumn{3}{|c||}{$\alpha=1.1$}&\multicolumn{3}{|c||}{$\alpha=1.5$}&\multicolumn{3}{|c|}{$\alpha=2.0$}\\
			\cline{3-11}			
			&Known$\dagger$&$^*$&+&Time$\ddagger$&$^*$&+&Time$\ddagger$&$^*$&+&Time$\ddagger$\\
		\hline
2&	1.13853E+09&	0.00&	680&	25.1&	0.00&	247&	27.5&	0.00&	109&	30.7\\
5&	1.88367E+08&	0.00&	1000&	27.0&	0.00&	994&	31.1&	0.00&	985&	35.1\\
10&	6.05127E+07&	0.00&	163&	34.7&	0.00&	125&	41.4&	0.00&	98&	48.0\\
15&	3.49362E+07&	0.00&	1&	41.9&	0.00&	2&	52.2&	0.00&	1&	62.7\\
20&	2.47108E+07&	0.00&	1&	46.1&	0.01&	1&	55.8&	0.29&	1&	73.8\\
25&	1.90899E+07&	0.31&	1&	63.6&	0.36&	1&	86.7&	0.47&	1&	122.5\\
			\hline							
			\multicolumn{11}{l}{$\dagger$ Best known solution including the results in this paper.}\\
			\multicolumn{11}{l}{$\ddagger$ Time in seconds for one run.}\\
			\multicolumn{11}{l}{$^*$ Percent above best known solution {\color{black} (relative error)}.}\\
			\multicolumn{11}{l}{+ Number of times (out of 1000) that the best solution found.}
		\end{tabular}
	\end{center}
\end{table}

\subsection{Summary of the Computational Experiments}

We tested the merging, construction and the separation (which is based on the construction) procedures as well as six algorithms run on R: H-W, Lloyd, and MacQ based on random starting solutions, and their variants marked with ``++" that require an additional code to design special starting solutions.

These algorithms were tested on ten data sets each with several instances for a total of 72 instances. Four small problems (36 instances) were basically solved to optimality, \citep[proven in ][]{A09}, by all approaches and no further comparisons are suggested in future research for these instances. The six medium and large problems are challenging, especially the {\color{black}kegg} instances. Twenty four instances, $k=10, 15, 20, 25$ for each problem, are compared below.

\begin{itemize}
\item Five new best known solutions were found by the algorithms proposed in this paper. Two of them were also found by the R implementations.
\item The best known solution was found by the algorithms proposed in this paper, within a standard number of runs, except for one instance {\color{black} tsplib1060}, $k=25$. \citet{BODX15,PABM19} report a solution of 6.06700E+08 while our best solution (found by the merging procedure using $\alpha=2$ run 1,000 times) is 6.06737E+08 which is 0.006\% higher. However, when we ran the merging process  10,000 times with $\alpha=2$ (required less than 7.5 minutes), we got a solution of 6.06607E+08, which is an improvement of the best known solution by 0.015\%.
\item The best result of the 6 R algorithms failed to obtain the best known solution in 9 of the 24 medium and large instances with $k\ge 10$ (see Table \ref{Rlrg}). In several cases, the deviation from the best known solution was large.
\end{itemize}

	\section{\label{sec5}Conclusions} 
	
Three new algorithms for generating starting solutions for the clustering problem and a new improvement algorithm are proposed. We extensively tested these algorithms on ten widely researched data sets with varying number of clusters for a total of 72 instances. Forty eight of these instances are relatively easy and all approaches, including standard approaches implemented in R, found the best known solutions.  Twenty four relatively large instances are more challenging. Five improved best known solutions were found by the three proposed algorithms and two of them were matched by the R procedures. The best known solutions were not found by the R implementations in 9 of the 24 instances.

It is well known that in the planar $p$-median location problem, as well as in other Operations Research problems, that starting with a good initial solution rather than a random one
significantly improves the final solution obtained with improvement algorithms.
This turns out to be true for the minimum sum of squared Euclidean distances clustering problem as well. The main contribution of the paper is finding good starting solutions for this clustering problem.

{\color{black} It would be interesting as a topic for future research to examine the performance of meta-heuristics using good starting solutions, such as the ones developed here, compared with random  starting solutions. This should lead to multi-start meta-heuristic based algorithms designed with higher intensifications, that are more effective then current approaches. It is possible that when better starting solutions are used, fewer multi-start replications are required to yield the same quality solutions.}	

\bigskip

\noindent{\bf Acknowledgment:}	The authors would like to thank professor Adil Bagirov from the Federation University of Australia for providing the source data and detailed results of the experiments reported in \citet{BODX15}.

	\renewcommand{\baselinestretch}{1}
	\renewcommand{\arraystretch}{1}
	\large
	\normalsize
	
	\bibliographystyle{apalike}

\begin{thebibliography}{}
	
{\color{black}{	\bibitem[Aloise, 2009]{A09}
	Aloise, D. (2009).
	\newblock {\em Exact algorithms for minimum sum-of-squares clustering}.
	\newblock PhD thesis, Ecole Polytechnique, Montreal, Canada.
	\newblock {ISBN}:978-0-494-53792-3.}}
	
{\color{black}{		\bibitem[Aloise et~al., 2012]{AHL12}
	Aloise, D., Hansen, P., and Liberti, L. (2012).
	\newblock An improved column generation algorithm for minimum sum-of-squares
	clustering.
	\newblock {\em Mathematical Programming}, 131:195--220.}}
	
	\bibitem[Arthur and Vassilvitskii, 2007]{David2007}
	Arthur, D. and Vassilvitskii, S. (2007).
	\newblock k-means++: The advantages of careful seeding.
	\newblock In {\em Proceedings of the eighteenth annual ACM-SIAM symposium on
		Discrete algorithms}, pages 1027--1035. Society for Industrial and Applied
	Mathematics.
	
	\bibitem[Bagirov et~al., 2015]{BODX15}
	Bagirov, A.~M., Ordin, B., Ozturk, G., and Xavier, A.~E. (2015).
	\newblock An incremental clustering algorithm based on hyperbolic smoothing.
	\newblock {\em Computational Optimization and Applications}, 61:219--241.
	
	\bibitem[Bahmani et~al., 2012]{BMV12}
	Bahmani, B., Moseley, B., Vattani, A., Kumar, R., and Vassilvitskii, S. (2012).
	\newblock Scalable k-means++.
	\newblock {\em Proceedings of the VLDB Endowment}, 5:622--633.
	
	\bibitem[Brimberg and Drezner, 2013]{BD12}
	Brimberg, J. and Drezner, Z. (2013).
	\newblock A new heuristic for solving the $p$-median problem in the plane.
	\newblock {\em Computers \& Operations Research}, 40:427--437.
	
	\bibitem[Brimberg and Drezner, 2019]{BD19}
	Brimberg, J. and Drezner, Z. (2019).
	\newblock Solving multiple facilities location problems with separated
	clusters.
	\newblock {\em Operations Research Letters}, 47:386--390.
	
	\bibitem[Brimberg et~al., 2012]{BDMS12a}
	Brimberg, J., Drezner, Z., Mladenovic, N., and Salhi, S. (2012).
	\newblock Generating good starting solutions for the $p$-median problem in the
	plane.
	\newblock {\em Electronic Notes in Discrete Mathematics}, 39:225--232.
	
	\bibitem[Brimberg et~al., 2000]{BHMT00}
	Brimberg, J., Hansen, P., Mladenovi{\'c}, N., and Taillard, E. (2000).
	\newblock Improvements and comparison of heuristics for solving the
	uncapacitated multisource {W}eber problem.
	\newblock {\em Operations Research}, 48:444--460.
	
{\color{black}	\bibitem[Brimberg et~al., 2017]{BMT17}
	Brimberg, J., Mladenovi{\'c}, N., Todosijevi{\'c}, R., and Uro{\v{s}}evi{\'c},
	D. (2017).
	\newblock Less is more: solving the max-mean diversity problem with variable
	neighborhood search.
	\newblock {\em Information Sciences}, 382:179--200.}
	
	\bibitem[Cooper, 1963]{Co63}
	Cooper, L. (1963).
	\newblock Location-allocation problems.
	\newblock {\em Operations Research}, 11:331--343.
	
	\bibitem[Cooper, 1964]{Co64}
	Cooper, L. (1964).
	\newblock Heuristic methods for location-allocation problems.
	\newblock {\em {SIAM} Review}, 6:37--53.
	
	\bibitem[Daskin, 1995]{Das95}
	Daskin, M.~S. (1995).
	\newblock {\em Network and Discrete Location: Models, Algorithms, and
		Applications}.
	\newblock John Wiley \& Sons, New York.
	
	\bibitem[Daskin and Maass, 2015]{DM15}
	Daskin, M.~S. and Maass, K.~L. (2015).
	\newblock The p-median problem.
	\newblock In Laporte, G., Nickel, S., and da~Gama, F.~S., editors, {\em
		Location science}, pages 21--45. Springer.
	
	\bibitem[Drezner, 2010]{Dr10b}
	Drezner, Z. (2010).
	\newblock Random selection from a stream of events.
	\newblock {\em Communications of the {ACM}}, 53:158--159.
	
	\bibitem[Drezner, 2015]{Dr13}
	Drezner, Z. (2015).
	\newblock The fortified {W}eiszfeld algorithm for solving the {W}eber problem.
	\newblock {\em {IMA} Journal of Management Mathematics}, 26:1--9.
	
	\bibitem[Drezner et~al., 2016]{DBMS13}
	Drezner, Z., Brimberg, J., Salhi, S., and Mladenovi{\'c}, N. (2016).
	\newblock New local searches for solving the multi-source {W}eber problem.
	\newblock {\em Annals of Operations Research}, 246:181--203.
	
	\bibitem[Feo and Resende, 1995]{FR95}
	Feo, T.~A. and Resende, M.~G. (1995).
	\newblock Greedy randomized adaptive search procedures.
	\newblock {\em Journal of global optimization}, 6:109--133.
	
	\bibitem[Forgy, 1965]{Forgy1965}
	Forgy, E.~W. (1965).
	\newblock Cluster analysis of multivariate data: efficiency versus
	interpretability of classifications.
	\newblock {\em Biometrics}, 21:768--769.
	
{\color{black}{			\bibitem[Gribel and Vidal, 2019]{GV19}
	Gribel, D. and Vidal, T. (2019).
	\newblock {HG}-means: A scalable hybrid genetic algorithm for minimum
	sum-of-squares clustering.
	\newblock {\em Pattern Recognition}, 88:569--583.}}
	
	\bibitem[Hartigan and Wong, 1979]{HW79}
	Hartigan, J. and Wong, M. (1979).
	\newblock Algorithm {AS} 136: A k-means clustering algorithm.
	\newblock {\em Journal of the Royal Statistical Society. Series C (Applied
		Statistics)}, 28:100--108.
	
{\color{black}	\bibitem[Kalczynski et~al., 2020]{KGD19}
	Kalczynski, P., Goldstein, Z., and Drezner, Z. (2020).
	\newblock Partitioning items into mutually exclusive groups.
	\newblock arXiv:2002.11536 [math.OC]  .}
	
	\bibitem[Kariv and Hakimi, 1979]{KH79med}
	Kariv, O. and Hakimi, S.~L. (1979).
	\newblock An algorithmic approach to network location problems. {II}: The
	$p$-medians.
	\newblock {\em {SIAM} Journal on Applied Mathematics}, 37:539--560.
	
	\bibitem[Kuenne and Soland, 1972]{KS72}
	Kuenne, R.~E. and Soland, R.~M. (1972).
	\newblock Exact and approximate solutions to the multisource {W}eber problem.
	\newblock {\em Mathematical Programming}, 3:193--209.
	
{\color{black}{			\bibitem[Lloyd, 1982]{L57}
	Lloyd, S. (1982).
	\newblock Least squares quantization in {PCM}.
	\newblock {\em IEEE transactions on information theory}, 28:129--137.}}
	
	\bibitem[MacQueen, 1967]{M67a}
	MacQueen, J. (1967).
	\newblock Some methods for classification and analysis of multivariate
	observations.
	\newblock In {\em Proceedings of the fifth Berkeley symposium on mathematical
		statistics and probability}, volume~1, pages 281--297. Oakland, CA, USA.
{\color{black}\bibitem[Mladenovi{\'c} et~al., 2016]{MTU16}
Mladenovi{\'c}, N., Todosijevi{\'c}, R., and Uro{\v{s}}evi{\'c}, D. (2016).
\newblock Less is more: basic variable neighborhood search for minimum
differential dispersion problem.
\newblock {\em Information Sciences}, 326:160--171.}
	
	\bibitem[Okabe et~al., 2000]{OKSC00}
	Okabe, A., Boots, B., Sugihara, K., and Chiu, S.~N. (2000).
	\newblock {\em Spatial Tessellations: Concepts and Applications of {V}oronoi
		Diagrams}.
	\newblock Wiley Series in Probability and Statistics. John Wiley.
	
{\color{black}{			\bibitem[Pereira et~al., 2018]{PABM19}
	Pereira, T., Aloise, D., Brimberg, J., and Mladenovi{\'c}, N. (2018).
	\newblock Review of basic local searches for solving the minimum sum-of-squares
	clustering problem.
	\newblock In {\em Open Problems in Optimization and Data Analysis}, pages
	249--270. Springer.}}
	
{\color{black}{			\bibitem[Sp\"ath, 1985]{S85}
	Sp\"ath, H. (1985).
	\newblock {\em The cluster dissection and analysis theory fortran programs
		examples}.
	\newblock Prentice-Hall, Inc.}}
	
	\bibitem[Vorono{\"\i}, 1908]{V08}
	Vorono{\"\i}, G. (1908).
	\newblock Nouvelles applications des param{\`e}tres continus {\`a} la
	th{\'e}orie des formes quadratiques. deuxi{\`e}me m{\'e}moire. recherches sur
	les parall{\'e}llo{\`e}dres primitifs.
	\newblock {\em Journal f{\"u}r die reine und angewandte Mathematik},
	134:198--287.
	
{\color{black}{			\bibitem[Weiszfeld, 1937]{W36}
	Weiszfeld, E. (1937).
	\newblock Sur le point pour lequel la somme des distances de n points
	donn{\'e}s est minimum.
	\newblock {\em Tohoku Mathematical Journal, First Series}, 43:355--386.}}
	
\end{thebibliography}

\end{document}